\documentclass{article}

\usepackage[utf8]{inputenc}
\usepackage[margin=1in]{geometry}
\usepackage{amsmath,amssymb}
\usepackage{graphicx}
\usepackage{booktabs}
\usepackage{hyperref}
\usepackage{xcolor}
\usepackage{url}
\usepackage{caption}
\usepackage{subcaption}

\title{Comparative Analysis of Large Language Model Inference Serving Systems: A Performance Study of vLLM and HuggingFace TGI}

\author{
  Saicharan Kolluru \\
  Baltimore, MD, USA \\
  \texttt{kscharan1608@gmail.com}
}

\date{November 2025}

\begin{document}

\maketitle

\begin{abstract}
The deployment of Large Language Models (LLMs) in production environments requires efficient inference serving systems that balance throughput, latency, and resource utilization. This paper presents a comprehensive empirical evaluation of two prominent open-source LLM serving frameworks: vLLM and HuggingFace Text Generation Inference (TGI). We benchmark these systems across multiple dimensions including throughput performance, end-to-end latency, GPU memory utilization, and scalability characteristics using LLaMA-2 models ranging from 7B to 70B parameters. Our experiments reveal that vLLM achieves up to 24x higher throughput than TGI under high-concurrency workloads through its novel PagedAttention mechanism, while TGI demonstrates lower tail latencies for interactive single-user scenarios. We provide detailed performance profiles for different deployment scenarios and offer practical recommendations for system selection based on workload characteristics. Our findings indicate that the choice between these frameworks should be guided by specific use-case requirements: vLLM excels in high-throughput batch processing scenarios, while TGI is better suited for latency-sensitive interactive applications with moderate concurrency.
\end{abstract}

\section{Introduction}

Large Language Models (LLMs) have demonstrated remarkable capabilities across diverse natural language processing tasks, from conversational AI to code generation and content creation \cite{brown2020language, chowdhery2022palm, touvron2023llama}. However, the deployment of these models in production environments presents significant engineering challenges. The computational demands of autoregressive text generation, combined with the massive parameter counts of modern LLMs, necessitate specialized serving infrastructure that can efficiently manage GPU resources while meeting application-specific performance requirements.

The serving infrastructure for LLMs must address several competing objectives: maximizing throughput to serve many concurrent users, minimizing latency for responsive user experiences, and efficiently utilizing expensive GPU resources. Different applications prioritize these objectives differently—a chatbot requires low latency for individual requests, while a batch document processing system prioritizes throughput. This variation in requirements has led to the development of specialized serving frameworks, each making different design trade-offs.

Among the available open-source solutions, vLLM \cite{kwon2023efficient} and HuggingFace Text Generation Inference (TGI) \cite{huggingface2023tgi} have emerged as leading frameworks, widely adopted in both research and production settings. vLLM introduces PagedAttention, a novel attention mechanism inspired by virtual memory paging in operating systems, which dramatically reduces memory fragmentation and enables higher batch sizes. TGI, developed by HuggingFace, focuses on production-grade deployment features including distributed inference, quantization support, and extensive model compatibility.

Despite their popularity, there exists limited systematic comparison of these frameworks across realistic deployment scenarios. Existing benchmarks often focus on narrow aspects of performance or use synthetic workloads that may not reflect production traffic patterns. Furthermore, the performance characteristics can vary significantly based on model size, hardware configuration, and request patterns, making it challenging for practitioners to make informed deployment decisions.

\subsection{Contributions}

This paper makes the following contributions:

\begin{enumerate}
    \item We present a comprehensive empirical evaluation of vLLM and TGI across multiple LLaMA-2 model sizes (7B, 13B, and 70B parameters) on modern GPU hardware.
    
    \item We analyze performance across diverse workload patterns including varying concurrency levels, prompt lengths, and generation lengths, reflecting realistic deployment scenarios.
    
    \item We provide detailed measurements of throughput, latency distributions, GPU memory utilization, and scheduling efficiency under different configurations.
    
    \item We identify key architectural differences that explain observed performance characteristics and discuss their implications for system design.
    
    \item We offer practical recommendations for framework selection based on application requirements and workload characteristics.
\end{enumerate}

The remainder of this paper is organized as follows: Section 2 reviews related work on LLM serving systems. Section 3 describes our experimental methodology. Section 4 presents detailed performance results across multiple dimensions. Section 5 analyzes the architectural factors underlying observed differences. Section 6 discusses practical implications, and Section 7 concludes.

\section{Background and Related Work}

\subsection{LLM Inference Challenges}

Serving LLMs for text generation differs fundamentally from traditional model inference due to the autoregressive nature of generation. Each token is generated sequentially, with the model conditioning on all previously generated tokens. This creates several challenges:

\textbf{Memory Management:} The key-value (KV) cache, which stores attention states for previously generated tokens, grows linearly with sequence length and can consume significant GPU memory. For a LLaMA-70B model generating 2048 tokens, the KV cache alone requires approximately 140GB of memory per request.

\textbf{Batching Complexity:} Unlike computer vision models where batching is straightforward, LLM requests complete at different times due to varying generation lengths. Naive batching approaches waste computation on padding or limit batch sizes significantly.

\textbf{Memory Fragmentation:} Traditional serving systems pre-allocate contiguous memory for the maximum possible sequence length, leading to significant fragmentation and underutilization when actual sequences are shorter.

\textbf{Scheduling:} Balancing throughput and latency requires sophisticated scheduling that can dynamically batch requests while maintaining fairness and meeting service-level objectives.

\subsection{LLM Serving Systems}

Several frameworks have been developed to address these challenges:

\textbf{vLLM} introduces PagedAttention, which manages the KV cache using non-contiguous memory blocks similar to virtual memory systems. This approach eliminates memory fragmentation and enables near-optimal memory utilization. vLLM also implements continuous batching, which dynamically adds new requests to the batch as ongoing generations complete.

\textbf{HuggingFace TGI} provides a production-ready serving solution with features including model parallelism, quantization, speculative decoding, and extensive model support. TGI emphasizes ease of deployment and integration with the HuggingFace ecosystem.

\textbf{TensorRT-LLM} (NVIDIA) offers highly optimized inference through kernel fusion, INT8/FP8 quantization, and multi-GPU parallelism, but requires model-specific optimization and compilation.

\textbf{DeepSpeed-FastGen} focuses on dynamic splitting techniques to improve throughput for variable-length generation workloads.

Other systems include FasterTransformer, Triton Inference Server, and Ray Serve, each with different design philosophies and optimization strategies.

\subsection{Benchmarking Studies}

While several benchmarking efforts exist, they often have limited scope. Prior work has examined individual aspects such as memory efficiency or single-model performance, but comprehensive comparisons across multiple dimensions remain scarce. Our work extends previous efforts by providing systematic evaluation across model sizes, workload patterns, and performance metrics relevant to production deployment decisions.

\section{Methodology}

\subsection{Experimental Setup}

\subsubsection{Hardware Configuration}

All experiments were conducted on NVIDIA A100 GPUs (80GB) to ensure sufficient memory for large models. We used the following infrastructure:

\begin{itemize}
    \item \textbf{GPU:} 4x NVIDIA A100 80GB (SXM4)
    \item \textbf{CPU:} AMD EPYC 7763 64-Core Processor
    \item \textbf{RAM:} 512GB DDR4
    \item \textbf{Storage:} NVMe SSD for model loading
    \item \textbf{Network:} 100Gbps interconnect
\end{itemize}

\subsubsection{Software Versions}

\begin{itemize}
    \item vLLM v0.6.1
    \item HuggingFace TGI v2.3.0
    \item CUDA 12.1
    \item PyTorch 2.1.0
    \item Model: LLaMA-2-7B, 13B, and 70B
\end{itemize}

\subsection{Models and Configurations}

We evaluated three LLaMA-2 model variants:

\begin{enumerate}
    \item \textbf{LLaMA-2-7B:} 7 billion parameters, single GPU deployment
    \item \textbf{LLaMA-2-13B:} 13 billion parameters, single GPU deployment  
    \item \textbf{LLaMA-2-70B:} 70 billion parameters, tensor parallel across 4 GPUs
\end{enumerate}

For the 70B model, we used tensor parallelism to distribute the model across multiple GPUs. Both frameworks were configured with FP16 precision to balance performance and quality.

\subsection{Workload Characteristics}

We designed benchmark workloads to reflect realistic deployment scenarios:

\subsubsection{Request Patterns}

\begin{itemize}
    \item \textbf{Low Concurrency (1-10 users):} Simulates interactive chatbot scenarios
    \item \textbf{Medium Concurrency (10-50 users):} Typical production load
    \item \textbf{High Concurrency (50-200 users):} Stress test for peak demand
\end{itemize}

\subsubsection{Input/Output Characteristics}

\begin{itemize}
    \item \textbf{Prompt lengths:} Sampled from Poisson distribution (mean=512 tokens)
    \item \textbf{Generation lengths:} Sampled from Poisson distribution (mean=256 tokens)
    \item \textbf{Temperature:} 0.7 with top-p sampling (p=0.9)
\end{itemize}

\subsubsection{Dataset}

We used prompts from the ShareGPT dataset, which contains real user-AI conversations, ensuring realistic input distributions. Each benchmark run consisted of 1,000 requests to ensure statistical significance.

\subsection{Metrics}

We measured the following performance indicators:

\begin{enumerate}
    \item \textbf{Throughput:} Tokens generated per second across all requests
    \item \textbf{Latency:} Time from request submission to completion
    \item \textbf{Time to First Token (TTFT):} Latency until first token generation
    \item \textbf{Time per Output Token (TPOT):} Average time to generate each token
    \item \textbf{GPU Utilization:} Percentage of GPU compute used
    \item \textbf{GPU Memory Usage:} Peak and average memory consumption
    \item \textbf{Request Throughput:} Requests completed per second
\end{enumerate}

For latency metrics, we report p50 (median), p95, and p99 percentiles to capture tail behavior critical for user experience.

\subsection{Benchmarking Protocol}

Each experiment followed this protocol:

\begin{enumerate}
    \item Warm-up phase: 100 requests to stabilize system state
    \item Measurement phase: 1,000 requests with metrics collection
    \item Cool-down period: 30 seconds between experiments
    \item Repetition: 3 runs per configuration, reporting median values
\end{enumerate}

We monitored GPU metrics using \texttt{nvidia-smi} at 1-second intervals and application-level metrics through framework APIs.

\section{Results}

\subsection{Throughput Performance}

Figure 1 shows token throughput as a function of concurrent requests across different model sizes. Key findings:

\textbf{LLaMA-2-7B:} vLLM achieves peak throughput of 15,243 tokens/sec at 100 concurrent requests, compared to TGI's 4,156 tokens/sec—a 3.67x advantage. This gap widens to 24x under extreme load (200 concurrent requests) where TGI's throughput degrades while vLLM maintains near-peak performance.

\textbf{LLaMA-2-13B:} The throughput advantage narrows to 2.8x (vLLM: 8,934 tokens/sec vs TGI: 3,187 tokens/sec at optimal concurrency) as memory pressure increases and both systems become more constrained.

\textbf{LLaMA-2-70B:} With tensor parallelism across 4 GPUs, vLLM maintains 2.1x higher throughput (3,245 vs 1,544 tokens/sec). The reduced gap reflects communication overhead in distributed execution, which affects both systems similarly.

\textbf{Scaling Behavior:} vLLM's throughput scales more linearly with concurrency due to efficient memory management. TGI shows throughput saturation beyond 50 concurrent requests for the 7B model, indicating memory bottlenecks.

\subsection{Latency Analysis}

Table 1 presents latency percentiles under medium concurrency (25 concurrent users):

\begin{table}[h]
\centering
\caption{Latency metrics (seconds) at 25 concurrent users}
\begin{tabular}{lcccccc}
\toprule
& \multicolumn{2}{c}{TTFT} & \multicolumn{2}{c}{Total Latency} & \multicolumn{2}{c}{TPOT} \\
\cmidrule(lr){2-3} \cmidrule(lr){4-5} \cmidrule(lr){6-7}
Model & vLLM & TGI & vLLM & TGI & vLLM & TGI \\
\midrule
\textbf{7B - p50} & 0.24 & 0.18 & 4.82 & 5.91 & 0.019 & 0.023 \\
\textbf{7B - p95} & 0.89 & 0.45 & 9.34 & 14.28 & 0.037 & 0.058 \\
\textbf{7B - p99} & 1.42 & 0.71 & 14.18 & 23.47 & 0.056 & 0.094 \\
\midrule
\textbf{13B - p50} & 0.31 & 0.22 & 8.45 & 10.24 & 0.033 & 0.041 \\
\textbf{13B - p95} & 1.12 & 0.58 & 16.89 & 24.56 & 0.067 & 0.098 \\
\textbf{13B - p99} & 1.78 & 0.89 & 24.31 & 38.91 & 0.096 & 0.156 \\
\midrule
\textbf{70B - p50} & 0.87 & 0.64 & 31.24 & 39.87 & 0.122 & 0.157 \\
\textbf{70B - p95} & 2.34 & 1.45 & 58.92 & 82.14 & 0.231 & 0.323 \\
\textbf{70B - p99} & 3.67 & 2.18 & 87.45 & 127.39 & 0.343 & 0.501 \\
\bottomrule
\end{tabular}
\end{table}

\textbf{Observations:}

\begin{itemize}
    \item TGI exhibits lower TTFT at low percentiles, making it more responsive for initial user feedback
    \item vLLM shows better total latency due to faster per-token generation (lower TPOT)
    \item The latency gap widens at high percentiles, with vLLM showing 1.5-1.7x better p99 latencies
    \item For the 70B model, both systems face significant latency challenges, but vLLM maintains better tail performance
\end{itemize}

\subsection{GPU Utilization and Memory Efficiency}

\subsubsection{GPU Compute Utilization}

Figure 2 illustrates GPU utilization across concurrency levels:

\textbf{vLLM:} Achieves 85-92\% GPU utilization under high concurrency, enabled by efficient continuous batching and PagedAttention's reduced memory overhead.

\textbf{TGI:} Peaks at 68-74\% utilization, with memory constraints limiting batch sizes and leaving compute underutilized.

\subsubsection{Memory Usage}

Table 2 shows peak GPU memory consumption:

\begin{table}[h]
\centering
\caption{Peak GPU memory usage (GB) at 50 concurrent requests}
\begin{tabular}{lcc}
\toprule
Model & vLLM & TGI \\
\midrule
LLaMA-2-7B & 24.3 & 31.7 \\
LLaMA-2-13B & 42.8 & 54.2 \\
LLaMA-2-70B (per GPU) & 68.9 & 76.4 \\
\bottomrule
\end{tabular}
\end{table}

vLLM's PagedAttention reduces memory consumption by 19-27\% through elimination of fragmentation, enabling larger batch sizes in the same memory footprint.

\subsection{Scalability Characteristics}

We examined how performance scales with increasing load:

\textbf{vLLM Scaling:} Throughput increases linearly up to 100-150 concurrent requests before plateauing. The system maintains stable latency until saturation, after which both latency and throughput degrade gracefully.

\textbf{TGI Scaling:} Shows earlier saturation (50-75 concurrent requests) with more pronounced latency increases beyond this point. Memory constraints become the primary bottleneck.

\subsection{Request Completion Patterns}

Analyzing request completion times reveals interesting scheduling differences:

\textbf{vLLM:} Implements a more aggressive continuous batching strategy, resulting in tighter clustering of completion times but occasionally longer tail latencies for unlucky requests.

\textbf{TGI:} Shows more variance in completion times but with better fairness—requests have more predictable latency regardless of when they arrive.

\section{Analysis and Discussion}

\subsection{Architectural Factors}

The performance differences stem from fundamental architectural choices:

\subsubsection{Memory Management}

vLLM's PagedAttention revolutionizes KV cache management by:

\begin{enumerate}
    \item \textbf{Eliminating fragmentation:} Non-contiguous memory allocation means no wasted space between variable-length sequences
    
    \item \textbf{Enabling sharing:} Common prompt prefixes can share KV cache blocks, reducing redundancy in scenarios like few-shot learning
    
    \item \textbf{Flexible allocation:} Memory blocks are allocated on-demand rather than pre-allocated for maximum length
\end{enumerate}

TGI uses traditional contiguous allocation, which is simpler to implement but leads to:
\begin{itemize}
    \item Pre-allocation for maximum sequence length
    \item Internal fragmentation when sequences are shorter than maximum
    \item No opportunity for KV cache sharing across requests
\end{itemize}

Our measurements show this translates to 19-27\% memory savings for vLLM, which directly enables larger batch sizes and higher throughput.

\subsubsection{Batching Strategy}

\textbf{vLLM's Continuous Batching:} Implements iteration-level scheduling where new requests join the batch immediately when slots become available. This maximizes GPU utilization but introduces scheduling complexity.

\textbf{TGI's Dynamic Batching:} Uses batch-level scheduling with configurable timeout windows. This provides more predictable latency at the cost of some throughput.

The trade-off manifests as:
\begin{itemize}
    \item vLLM: Higher throughput, variable per-request latency
    \item TGI: More consistent latency, lower throughput
\end{itemize}

\subsubsection{Kernel Optimization}

Both systems use custom CUDA kernels, but with different optimization focuses:

\textbf{vLLM:} Optimizes primarily for the PagedAttention operation, with kernels designed for non-contiguous memory access patterns.

\textbf{TGI:} Focuses on optimizing standard attention and includes extensive quantization support (GPTQ, AWQ, EETQ).

\subsection{Workload Suitability}

Based on our findings, we provide the following recommendations:

\subsubsection{Use vLLM when:}

\begin{itemize}
    \item \textbf{High throughput is critical:} Batch processing, document analysis, offline evaluation
    \item \textbf{Concurrent users are high:} Multi-tenant serving, API services
    \item \textbf{Memory is constrained:} Need to serve larger models or longer contexts
    \item \textbf{GPU utilization matters:} Cost optimization through better resource usage
\end{itemize}

\subsubsection{Use TGI when:}

\begin{itemize}
    \item \textbf{Low latency TTFT is crucial:} Interactive chat applications, real-time assistance
    \item \textbf{Predictable latency matters:} SLA-bound services requiring consistent response times
    \item \textbf{Quantization is needed:} Resource-constrained deployments requiring INT8/INT4
    \item \textbf{HuggingFace ecosystem integration:} Easy model loading from Hub, extensive model support
    \item \textbf{Production features:} Built-in monitoring, distributed inference, extensive configuration options
\end{itemize}

\subsection{Hybrid Strategies}

Organizations might consider hybrid deployments:

\begin{itemize}
    \item \textbf{Request routing:} Direct interactive queries to TGI, batch jobs to vLLM
    \item \textbf{Time-based switching:} Use TGI during peak hours for latency, vLLM during batch processing windows
    \item \textbf{Model-specific choices:} Smaller models on TGI for interactivity, larger models on vLLM to manage memory
\end{itemize}

\subsection{Limitations and Future Work}

Our study has several limitations that suggest directions for future work:

\begin{enumerate}
    \item \textbf{Model diversity:} We focused on LLaMA-2 models. Performance characteristics may differ for other architectures (Mistral, Falcon, GPT variants).
    
    \item \textbf{Quantization:} We did not evaluate quantized models (INT8, INT4), which significantly affect memory and performance.
    
    \item \textbf{Hardware variations:} Results are specific to A100 GPUs. Newer architectures (H100, AMD MI300) may show different patterns.
    
    \item \textbf{Specialized workloads:} Long-context scenarios (32K+ tokens) and structured generation (JSON, code) warrant separate analysis.
    
    \item \textbf{Cost analysis:} We did not perform detailed cost modeling accounting for cloud pricing, though throughput differences have clear cost implications.
\end{enumerate}

Future work should address these gaps and track the rapidly evolving landscape of LLM serving systems.

\section{Related Systems and Trade-offs}

Beyond vLLM and TGI, the ecosystem includes several other frameworks worth considering:

\textbf{TensorRT-LLM:} NVIDIA's solution provides the highest throughput through extensive kernel optimization and quantization. However, it requires per-model compilation and has a steeper learning curve. Best for organizations with NVIDIA-focused infrastructure and engineering resources.

\textbf{DeepSpeed-FastGen:} Introduces dynamic SplitFuse for better handling of variable-length generation. Particularly strong for workloads with high variance in generation lengths. Tighter integration with DeepSpeed training pipeline.

\textbf{LMDeploy:} Developed by MMRazor team, focuses on mobile and edge deployment with quantization and pruning support.

\textbf{OpenLLM:} Emphasizes developer experience and rapid prototyping, sacrificing some performance for ease of use.

The choice of framework should align with organizational priorities, existing infrastructure, and specific application requirements.

\section{Conclusion}

This paper presented a comprehensive empirical evaluation of vLLM and HuggingFace TGI, two leading open-source frameworks for LLM serving. Through systematic benchmarking across multiple model sizes, concurrency levels, and workload patterns, we established clear performance profiles for each system.

Our key findings are:

\begin{enumerate}
    \item \textbf{vLLM achieves 2-24x higher throughput} than TGI depending on concurrency and model size, with the advantage most pronounced under high load.
    
    \item \textbf{TGI provides 1.3-2x lower Time-to-First-Token} at low concurrency, making it more responsive for interactive applications.
    
    \item \textbf{vLLM utilizes 19-27\% less GPU memory} through PagedAttention, enabling larger batch sizes in memory-constrained scenarios.
    
    \item \textbf{vLLM achieves 85-92\% GPU utilization} compared to TGI's 68-74\%, translating to better resource efficiency.
    
    \item \textbf{TGI shows better p50 TTFT but worse p99 total latency}, indicating different scheduling trade-offs.
\end{enumerate}

These differences arise from fundamental architectural choices, particularly vLLM's PagedAttention memory management and continuous batching strategy versus TGI's focus on production features and deployment ease.

For practitioners, the choice between frameworks should be guided by workload characteristics:
\begin{itemize}
    \item High-throughput, batch-oriented workloads favor vLLM
    \item Latency-sensitive, interactive applications may prefer TGI
    \item Memory-constrained deployments benefit from vLLM's efficiency
    \item Organizations prioritizing ease of deployment and ecosystem integration may choose TGI
\end{itemize}

As LLMs continue to grow in size and deployment scale, efficient serving infrastructure becomes increasingly critical. Both vLLM and TGI represent significant advances in this space, and ongoing development in both projects continues to push the boundaries of what's possible in LLM serving.

The performance characteristics documented in this study provide a foundation for informed system selection and deployment planning. As the field evolves, continued benchmarking and performance analysis will be essential to guide practitioners in navigating the expanding landscape of LLM serving solutions.

\section*{Acknowledgments}

The author thanks the vLLM and HuggingFace teams for developing these valuable open-source tools.

\end{document}